\pdfoutput=1

\documentclass[11pt]{article}

\usepackage{multirow}
\usepackage{booktabs}
\usepackage{adjustbox}
\usepackage{array} 
\usepackage[section]{placeins}
\usepackage{bm}
\usepackage{float}
\usepackage{amsmath}
\usepackage{hyperref}

\usepackage{amssymb}
\newcommand{\mbf}[2][]{\bm{\mathrm{#2}}\ifx\relax#1\relax\else^{\text{#1}}\fi}

\usepackage[preprint]{acl}

\usepackage{times}
\usepackage{latexsym}

\usepackage[T1]{fontenc}

\usepackage[utf8]{inputenc}

\usepackage{microtype}

\usepackage{inconsolata}

\usepackage{graphicx}

\title{Adaptive Inner Speech-Text Alignment for LLM-based Speech Translation
}

\author{
 \textbf{Henglyu Liu\textsuperscript{\textdagger}},
 \textbf{Andong Chen\textsuperscript{\textdagger}},
 \textbf{Kehai Chen\textsuperscript{\textdagger}},
 \textbf{Xuefeng Bai\textsuperscript{\textdagger}},
\\
 \textbf{Meizhi Zhong\textsuperscript{\textdagger}},
 \textbf{Yuan Qiu\textsuperscript{\textdaggerdbl}},
 \textbf{Min Zhang\textsuperscript{\textdagger}},
\\
 \textsuperscript{\textdagger}Institute of Computing and Intelligence, Harbin Institute of Technology, Shenzhen, China
\\
 \textsuperscript{\textdaggerdbl}School of Computer Science and Engineering, Xi'an University of Technology
\\
 \texttt{23s151043@stu.hit.edu.cn, \{ands691119,meizhi.zhong.1999\}@gmail.com} \\
\texttt{ qiuyuan@xaut.edu.cn, \{chenkehai,baixuefeng,zhangmin2021\}@hit.edu.cn} \\
}

\begin{document}
\maketitle
\begin{abstract}
Recent advancement of large language models (LLMs) has led to significant breakthroughs across various tasks, laying the foundation for the development of LLM-based speech translation systems. Existing methods primarily focus on aligning inputs and outputs across modalities while overlooking deeper semantic alignment within model representations. To address this limitation, we propose an \textbf{A}daptive \textbf{I}nner \textbf{S}peech-\textbf{T}ext \textbf{A}lignment \textbf{(AI-STA)} method to bridge the modality gap by explicitly aligning speech and text representations at selected layers within LLMs. To achieve this, we leverage the optimal transport (OT) theory to quantify fine-grained representation discrepancies between speech and text. Furthermore, we utilize the cross-modal retrieval technique to identify the layers that are best suited for alignment and perform joint training on these layers. Experimental results on speech translation (ST) tasks demonstrate that \textbf{AI-STA} significantly improves the translation performance of large speech-text models (LSMs), outperforming previous state-of-the-art approaches. Our findings highlight the importance of inner-layer speech-text alignment in LLMs and provide new insights into enhancing cross-modal learning.
\end{abstract}

\section{Introduction}
The emergence of large language models (LLMs) \citep{brown2020language,touvron2023llama,anil2023palm,chiang2023vicuna} has achieved remarkable success across numerous natural language processing (NLP) tasks \citep{openai2024gpt4technicalreport} and various studies extend its generative capabilities to multimodal domains \citep{chen2023x,zhang2023video,li2023blip,rubenstein2023audiopalm,li2024llava}. The unprecedented capabilities of LLMs have laid the foundation for leveraging these models as the foundation for building powerful speech translation (ST) systems \cite{sethiya2024end}.

\begin{figure}[t!]
  \centering
  \includegraphics[width=0.85\columnwidth]{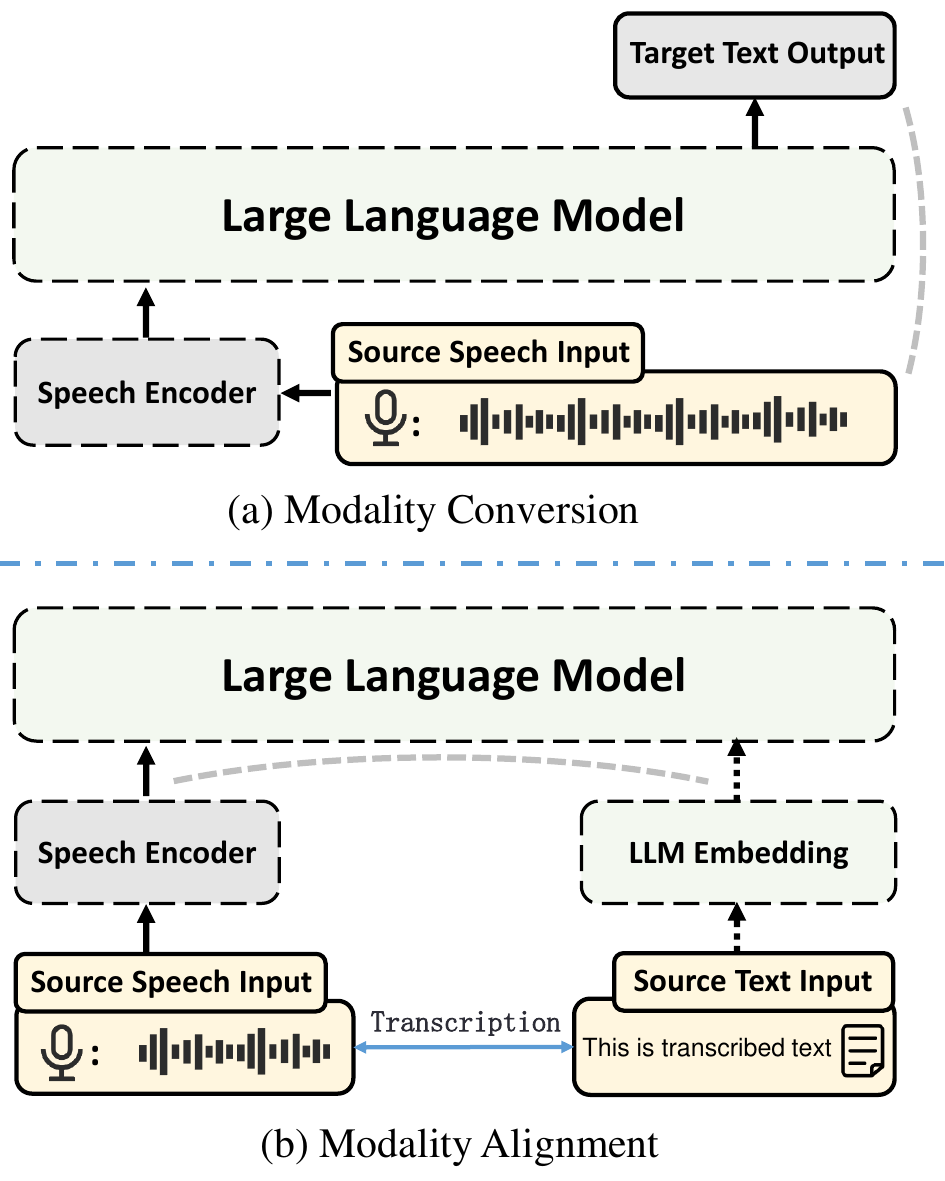}
  \caption{
  Different training paradigm: \textit{Modality Conversion} implicitly learns speech-text relationships from paired data, focusing on end-to-end mapping. While \textit{Modality Alignment} explicitly enforces semantic consistency by aligning representations through supervised objectives.
  }
  \label{fig:experiments}
\end{figure}

To equip text-based LLMs with speech capabilities, recent research has investigated multiple approaches for developing large speech-text models (LSMs). These methods include expanding text-based LLMs vocabulary and adopting large-scale speech-text pre-training \citep{rubenstein2023audiopalm, zhang2023speechgpt}, multi-task learning \cite{chu2023qwen}, curriculum learning \cite{hu2024wavllm}, constructing speech instruction fine-tuning datasets \citep{tang2023salmonn, wang2023blsp}. 
However, as illustrated in Figure~\ref{fig:experiments}, these approaches primarily concentrate on \textit{Modality Conversion} paradigm, which addresses the superficial relationship between the inputs and outputs of different modalities. It often leads to the neglect of the deeper semantic alignment, which is essential for ensuring that both speech and text embeddings convey equivalent meanings.

Motivated by these findings, we argue that \textit{Modality Alignment} paradigm which aligns speech and text representation is crucial for further improving the performance on ST tasks. 
To achieve this, we introduce optimal transport (OT) theory \cite{peyre2019computational} to capture the fine-grained representation differences between speech and text. Additionally, we propose a novel \textbf{A}daptive \textbf{I}nner \textbf{S}peech-\textbf{T}ext \textbf{A}lignment \textbf{(AI-STA)} method that dynamically selects specific layers within LLM to align speech and text representations. Experiments conducted on speech translation (ST) demonstrate that our method effectively improves the translation ability of LSM. 
Our main contributions are summarized as follows:
\begin{itemize}
\item We first explore the impact of the inner layer alignment between speech and text modalities in LLMs.
\item We propose an innovative adaptive speech-text alignment method to bridge the modality gap in specific selected layers and improve the performance of ST.
\item Extensive experiments demonstrate that \textbf{AI-STA} outperforms the previous state-of-the-art (SOTA) methods \cite{chu2024qwen2} on the CoVoST2 \cite{wang2021covost} dataset in two translation directions.
\end{itemize}
\section{Related Work}
\subsection{LLM-based Speech Translation}
LLM have demonstrated powerful semantic understanding and generation capabilities in translation tasks \citep{chen2024dual,zhang2024paying,chen2024make}, making them effective tools for solving ST tasks \cite{sethiya2024end}. Inspired by the aforementioned advantages, recent studies have leveraged the capabilities of LLMs to address a variety of downstream speech tasks. The prevailing method involves feeding discretized speech units into the LLM and expanding its vocabulary to enable understanding and generation of speech \citep{rubenstein2023audiopalm,zhang2023speechgpt,wang2024viola}. Another common approach is to connect a speech encoder to a backbone LLM, enabling effective processing of speech inputs \citealp{chu2023qwen,du2023lauragpt,chu2024qwen2,hu2024wavllm,fang2024llama}. These models support a wide range of multi-modal speech tasks while achieving comparable performance with task-specific ST models.

LST \cite{zhang2023tuning} employs Wav2vec 2.0 \cite{baevski2020wav2vec} as the fronted speech encoder and Llama 2 \cite{touvron2023llama} as LLM, achieving high performance on the MuST-C dataset \cite{di2019must}. \cite{huang2023speech} further incorporates the Chain-of-Thought (CoT) \cite{wei2022chain}, enabling a step-by-step approach using LLMs. LLaST \cite{chen2024llast} proposed a dual-LoRA optimization strategy rendering it a strong baseline for the CoVoST2 \cite{wang2021covost} in X->En translation direction.

However, these studies primarily focus on modality conversion, while the intrinsic semantic correlation between input speech and its transcript text has not been fully exploited. In this work, we emphasize the role of modality alignment between input speech and text and propose explicit supervision signals to guide the model in learning their underlying semantic relationships.

\subsection{Speech-Text Cross-Modality Alignment}
Cross-modal alignment aims to establish semantically consistent mappings between different modalities \cite{liang2022foundations}. Early cross-modal alignment methods for speech and text modalities were mostly implicit, relying on parameter-sharing encoding mechanisms and performing multi-task learning on paired speech-text data to align the speech and text spaces \citep{ao2021speecht5,bapna2021slam,tang2022unified}. Additionally, various approaches have been proposed to address modality differences by designing different loss functions and training objectives, such as connectionist temporal classification \citep{liu2020bridging,wang2020bridging,xu2021stacked}, contrastive learning \cite{ye2022cross,ouyang2022waco,fang2022stemm}, adversarial learning \cite{alinejad2020effectively}, and optimal transport \citep{zhou2023cmot,le2023pre,tsiamas2024pushing}. These methods have primarily been explored within the encoder-decoder architecture and applied to the embedding or encoder layers. 

Recent works have explored the alignment between speech and LLMs' text embeddings. For example, \cite{wang2024blsp} employ CFormer to address the speech-text length mismatch and introduce a KL-divergence loss to enhance the alignment of output distributions. \cite{nguyen2025spirit} conducts continuous training using mixed speech and text sequences, enabling the model to effectively learn cross-modal tasks.

In this study, we extend the cross-modal alignment to the decoder-only architecture and introduce a cross-modal retrieval task to investigate whether different hidden layers can effectively contribute to representation alignment.

\begin{figure}[t]
  \centering
  \includegraphics[width=\columnwidth]{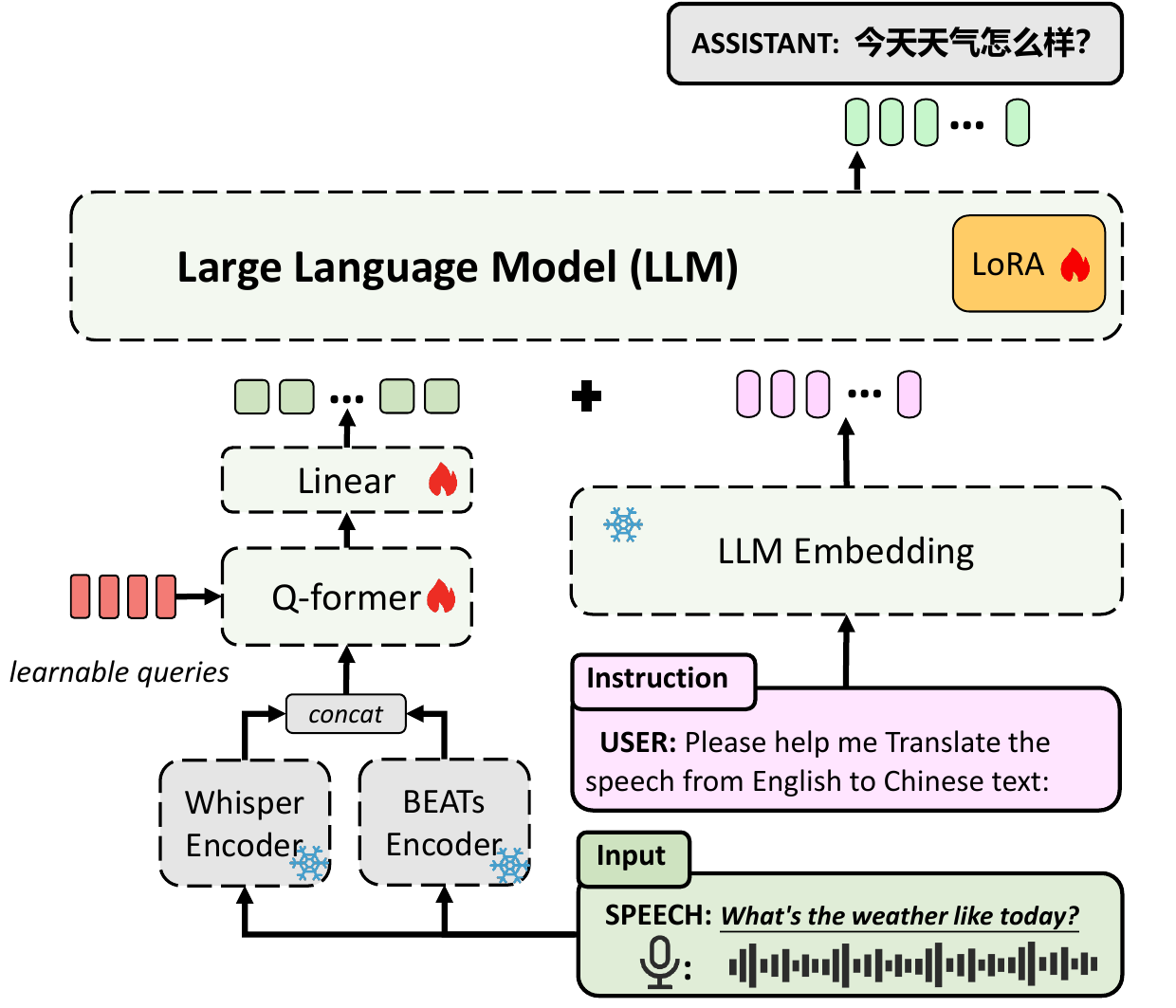}
  \caption{Model Architecture of Our LSM.}
  \label{fig:Method1}
\end{figure}

\section{Preliminary}
\subsection{Model Architecture}
As illustrated in Figure~\ref{fig:Method1}, the model architecture in this study is identical to SALMONN~\cite{tang2023salmonn}. We use the Whisper-Large encoder \cite{radford2023robust} as the speech encoder, BEATs \cite{chen2022beats} as the audio encoder, and employ Vicuna-13B-v1.1 \cite{chiang2023vicuna} or Qwne2-7B \cite{yang2024qwen2technicalreport} as the backbone LLM. Q-Former \cite{li2023blip} serves as the connection module followed by a linear layer to project speech representation to the text representation space. The output sequence integrated with the text instructions will be fed into the LLM with LoRA adapters \cite{hu2021lora} to generate the text response. LoRA as a widely used parameter-efficient fine-tuning method for LLM adaptation, introduces additional trainable parameters. The trainable parameters of our LSM include those from LoRA adapters, Q-Former, and the linear layer, while the backbone LLM and two encoders remain frozen during training.

\subsection{Optimal Transportation}
OT has recently been applied in ST, primarily for finding alignments between speech and text \cite{zhou2023cmot}, enhancing the effectiveness of speech pre-training \cite{le2023pre}, or integrating the speech encoder to the text space of the machine translation (MT) model \cite{tsiamas2024pushing}. While previous work has concentrated on encoder architectures, we extend this approach to a decoder-only architecture. To this end, we utilize OT to integrate the speech representation space into the text representation space within LLMs.

To align a speech representation $\mbf[s]{h} \in \mathbb{R}^{n \times d}$ with the text representation $\mbf[t]{h} \in \mathbb{R}^{m \times d}$, we minimize their Wasserstein loss \cite{frogner2015learning} using OT theory \citep{le2023pre, zhou2023cmot, tsiamas2024pushing}. We assume the mass of each position in the speech and text representations are two uniform probability distributions. The optimized objective is defined as:

 \begin{footnotesize}
\begin{equation}
\vspace{-0.1cm}
    \label{eq:Wass}
    \begin{split}
        & W_\delta = \underset{\mbf{Z}}{\text{min}} \sum_{i=1}^{n} \sum_{j=1}^m \mbf{Z}_{ij} \mbf{C}_{ij},
        \\
        \mbf{s.t.}~&\sum_{j=1}^m \mbf{Z}_{:,j} = \frac{1}{n},\sum_{i=1}^n \mbf{Z}_{i,:} = \frac{1}{m}.
    \end{split}
    \vspace{-0.1cm}
\end{equation}
\end{footnotesize}

The Wasserstein distance $W_\delta$ is defined as the minimum transportation cost of all possible transportation plans $\mbf{Z}$. $\mbf{C}$ represents a squared euclidean cost between two vectors, where $\mbf{C}_{ij} = {||h_i^s - h_j^t||}^2 $.

\section{Methodology}
\subsection{Speech Pre-training Stage} 
To enable LLM to initially comprehend speech inputs and mitigate the discrepancy between pre-trained parameters and randomly initialized parameters, we utilize extensive datasets focusing on recognition and annotation tasks. This phase aims to establish a foundational ability to handle spoken language rather than deeply understanding the textual content within these speeches.

\begin{figure*}[t]
  \centering
  \includegraphics[width=\linewidth]{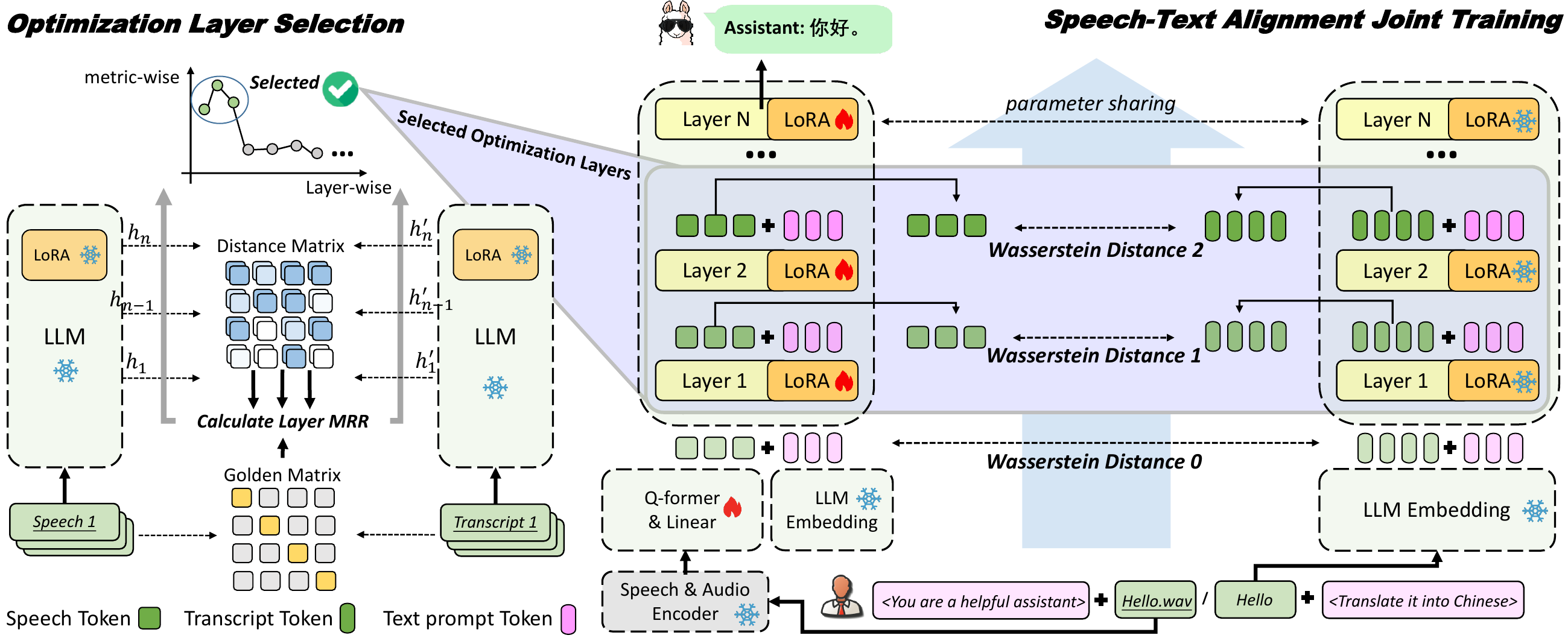}
  \caption{Overview of second and third stages of the proposed AI-STA. The \textbf{left} part first chose specific layers within the LLM according to its cross-modal retrieval ability. Then the \textbf{right} part obtains hidden states by separately forwarding speech or transcribed text concatenated with the same prompts and optimizes the LSM by combining alignment loss (computed via Wasserstein distance between hidden states) with cross-entropy loss.}
  \label{fig:Stage3}
  \vspace{-0.3cm}
\end{figure*}

Let \textbf{S} represent the speech input and \textbf{T} as its corresponding target text sentence. The speech encoder and audio encoder transform the speech input \textbf{S} into representations \bm{$R^{'}$}  and \bm{$R^{''}$}, respectively. Since both encoders have the same output frame rate of 50Hz, we finally get \textbf{R} by a frame-by-frame concatenation operation along the feature dimension. Then we use the window-level Q-Former \cite{tang2023salmonn} to segment \textbf{R} into \textit{L}-sized window representations and outputs textual tokens \bm{$E^{S}$}. The main training objective for the speech-text pair (\textbf{S}, \textbf{T}) is:

\begin{footnotesize}
\begin{equation}
\vspace{-0.1cm}
    \label{eq:training_objective}
    \begin{split}
        \hat{\theta} &= \mathop{\arg\min}\limits_{\theta} \left(-\log_{}{P \left( T \left| E^{S},I,\theta\right. \right)}\right),
                \\
        &= \arg\min_{\theta} \left(-\sum_{m=1}^M \log P(T_m | {T}_{<m}, {E^{S}}, {I}, \theta)\right),
    \end{split}
    \vspace{-0.1cm}
\end{equation}
\end{footnotesize}\\
where \textit{M} is the length of the target token, $T_m$ is the \textit{m}-th target token, and \textit{I} is the embedding of task instruction. We employed the standard causal language modeling loss as our training loss, which is designed to predict the subsequent token based on the previous token. We using the same prompt template as described by SALMONN \cite{tang2023salmonn} for Vicuna and Qwen2-Audio \cite{chu2024qwen2} for Qwen.

\subsection{Optimization Layer Selection Stage}
Figure~\ref{fig:Stage3} \textbf{left} part depicts the process of this stage. To determine the most suitable LLM layers for representation alignment, we conduct experiments after the speech pre-training stage. We randomly sample 1,000 parallel speech-text pairs from Librispeech test-clean set \cite{panayotov2015librispeech}. Each data pair, concatenated with the same instruction, is fed into the LLM to extract hidden states from all layers. Let $h^s_{i,l}$ denote the \textit{l}-th layer and \textit{i}-th sample speech representation and $h^t_{i,l}$ denote the corresponding text representation. We then compute the Wasserstein Distance \cite{frogner2015learning} for each speech-text pair, constructing a distance matrix that facilitated speech-to-text retrieval, as described by the following equation:

\begin{footnotesize}
\begin{equation}
\vspace{-0.1cm}
    \label{eq:wd}
    \begin{split}
        D_{i,j}^{(l)} = Wasserstein(h^s_{i,l}, h^t_{j,l}).
    \end{split}
    \vspace{-0.1cm}
\end{equation}
\end{footnotesize}

\begin{figure}[t]
  \includegraphics[width=\columnwidth]{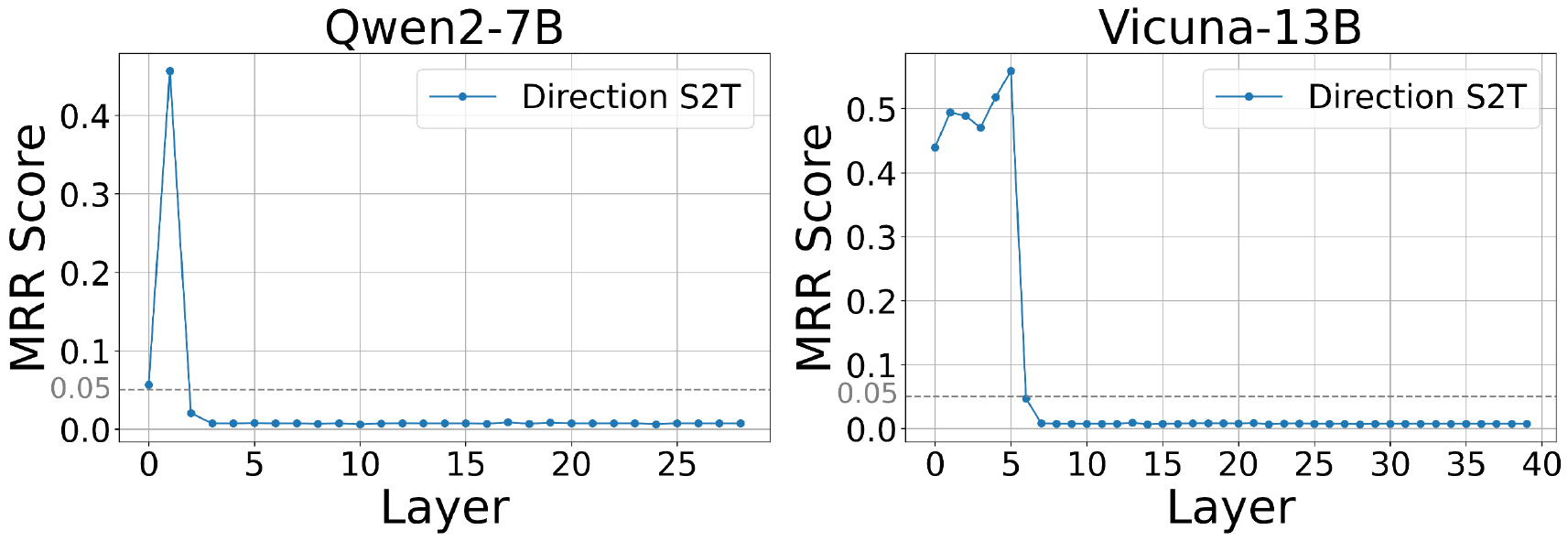}
  \caption{Layer-wise trends of average mean reciprocal rank (MRR) in two distinct backbone LLMs for speech-to-text retrieval evaluation.}
  \label{fig:Layer_Selection}
\end{figure}

Specifically, we rank the text samples according to their Wasserstein Distance from the speech samples and calculate the mean reciprocal rank (MRR) of the golden match across all 1,000 samples. The metrics are expressed as follows:

\begin{footnotesize}
\begin{equation}
\vspace{-0.1cm}
    \label{eq:mrr}
    \begin{split}
        {MRR}= \frac{1}{Q} \sum_{i=1}^{Q} \frac{1}{r_{i}},
    \end{split}
    \vspace{-0.1cm}
\end{equation}
\end{footnotesize}\\
where $Q$ is the number of speech-text pairs. For each speech utterance, define $r_{i}$ as the actual rank of the ground-truth paired text among all text samples. Subsequently, we compute the average MRR for each layer:

\begin{footnotesize}
\begin{equation}
\vspace{-0.1cm}
    \label{eq:selected_layer}
    \begin{split}
        \mathcal{I} = \{l | {MRR}^{(l)}>threshold, l \in [0,num\_layer]\},
    \end{split}
    \vspace{-0.1cm}
\end{equation}
\end{footnotesize}\\
where $num\_layer$ is determined by the LLM we used. The 0-th layer corresponds to the embedding layer. As shown in Figure~\ref{fig:Layer_Selection}, we observe that in the Vicuna-13B version, the MRR scores remain relatively high (above 0.5) from layer 0 (the embedding layer) to layer 5. However, starting from layer 6, the scores drop sharply, falling below 0.01. Similarly, in the Qwen2-7B version, layers 0 and 1 demonstrate higher scores but exhibit a steep decline in subsequent layers. This pattern indicates that the shallow layers of LLM play a crucial role in capturing the semantic properties of speech inputs. Based on empirical practice, a threshold value of 0.05 was set to select our optimal layers for optimization.

\subsection{Speech-Text Alignment Joint Training Stage}
As illustrated in Figure~\ref{fig:Stage3} \textbf{right} part, we fine-tune the model for downstream speech tasks while simultaneously training for speech-text alignment. Specifically, the transcribed text and instruction are concatenated and passed through the LLM, with their representation positions recorded to extract the corresponding text and speech representations. 
For the selected optimization layers, pairwise Wasserstein loss between these representations was minimized. The gradients from text representations do not contribute to the optimization process. Therefore, the final loss is defined as:

\begin{footnotesize}
\begin{equation}
\vspace{-0.1cm}
    \label{eq:loss}
    \begin{split}
        &\mathcal{L}_\text{CE} = -\log_{}{P \left( T \left| E^{S},I,\theta\right. \right)}, \\
        &\mathcal{L} = \alpha\mathcal{L}_\text{CE} + \sum_{l \in \mathcal{I}} \frac{1-\alpha}{|\mathcal{I}|} \mathcal{L}_\text{Wass}^{(l)},
    \end{split}
    \vspace{-0.1cm}
\end{equation}
\end{footnotesize}\\
where $\mathcal{L}_\text{Wass}^{(l)}$ is equivalent to the Wasserstein distance between speech and text representations in the \textit{l}-th layer, and $\alpha$ is a hyperparameter to balance the relative importance between two loss.

\section{Experiment}
\subsection{Training Data}
In the speech pre-training stage, we use LibriSpeech training set \cite{panayotov2015librispeech} and GigaSpeech M-set \cite{chen2021gigaspeech} for automatic speech recognition (ASR), as well as WavCaps \cite{mei2024wavcaps} (with audio clips longer than 180 seconds removed), AudioCaps \cite{kim2019audiocaps} and Clotho \cite{drossos2020clotho} dataset for automatic audio captioning (AAC). 

In the joint training stage, we chose the ST task for further training. CoVoST2 \cite{wang2021covost} is a large-scale multilingual dataset that supports translations between English and 15 other languages, as well as from 21 languages into English. To align with the previous method, we select two translation directions including English-Chinese and English-Japanese. All the datasets we used are listed in the Table~\ref{tab:training_dataset}.
\subsection{Training Setup}
Our model employs the encoder part of Whisper-Large-v2 \cite{radford2023robust} model as the speech encoder, the fine-tuned BEATs \cite{chen2022beats} encoder as the audio encoder, and a Vicuna-13B-v1.1 \cite{chiang2023vicuna} or a Qwen2-7B \cite{yang2024qwen2technicalreport} as the backbone LLM. In the Q-Former block, we set \textit{N = 1} for a single trainable query and use \textit{L = 17} which is approximately 0.33 seconds per window. The OT loss weight is empirically set to 0.01 based on practical experience. We freeze speech encoder, audio encoder, and LLM when training, leading 28 million (M) or 64M trainable parameters, depending on the scale of the backbone LLM parameters. Detailed training hyperparameters are available in Appendix~\ref{sec:Hyperparameters}.
\begin{table}[!t]
  \centering
  \renewcommand\arraystretch{1.0}
  \begin{tabular}{
    >{\centering\arraybackslash}p{0.05\textwidth}
    >{\centering\arraybackslash}p{0.18\textwidth}
    >{\centering\arraybackslash}p{0.06\textwidth}
    >{\centering\arraybackslash}p{0.09\textwidth}}
    \toprule
    \textbf{Task} & \textbf{Data Source} & \textbf{\#Hours} & \textbf{\#Samples} \\
    \hline
    
    \multirow{2}*{ASR} &  {LibriSpeech}    & {960}    & {280K}     \\
    &  {GigaSpeech M-set}    & {1000}    & {680K}     \\
    \hline
    \multirow{3}*{AAC} &  {WavCaps}    & {2800}    & {370K}     \\
    &  {AudioCaps}    & {-}    & {45K}     \\
    &  {Clotho}    & {-}    & {4K}     \\
    \hline
    \multirow{2}*{ST}  & {CoVoST2 En2Zh}    & {364}    & {289K}     \\
    & {CoVoST2 En2Ja}   & {364}    & {289K}     \\
    \bottomrule
  \end{tabular}
  \caption{Training data used in all stages.}
  \label{tab:training_dataset}
\end{table}

\subsection{Evaluation}
We evaluated the model using the CoVoST2 test set for English-Chinese and English-Japanese translations, employing SacreBLEU \cite{post2018call} score as the evaluation metric. Audio samples are all resampled to 16kHz in the experiments. 

\subsection{Baselines}
We compare our LSM and method with the following four baselines.

\textbf{SALMONN}~\cite{tang2023salmonn} integrates a pre-trained text-based LLM with a speech encoder and audio encoder to process audio inputs. It excels in tasks like speech recognition, translation, and music captioning while showcasing emergent abilities. 

\textbf{BLSP-KD}~\cite{wang2024blsp} leverages CFormer architecture to tackle the speech-text length discrepancy, while incorporating a KL-divergence loss mechanism to optimize output distribution alignment. It also introduces a partial LoRA strategy to facilitate efficient LLM fine-tuning with speech inputs.

\begin{table}[t]
  \centering
  \renewcommand\arraystretch{1.0}
  \begin{tabular}{
    p{0.2\textwidth}|
    >{\centering\arraybackslash}p{0.1\textwidth}|
    >{\centering\arraybackslash}p{0.1\textwidth}}
    \toprule
    \multirow{2}{*}{\textbf{Method}}
     & \multicolumn{2}{c}{\textbf{CoVoST2}}
    \\
    & En-Zh & En-Ja \\
    \hline
    \multicolumn{3}{c}{\textit{Baseline Models}} \\
    {SALMONN}        & {33.1}    & {22.7}   \\

    {BLSP-KD}                & {41.3}    & {21.3}   \\
    {Qwen-Audio}        & {41.5}    & {23.5}   \\

    {Qwen2-Audio}              & \underline{45.2}    & \underline{28.6}  \\
    \hline
    \multicolumn{3}{c}{\textit{Our LSM with Vicuna-13B-v1.1}} \\
    {base}    & {36.5}    & {29.8}   \\
    {w/ AI-STA}    & {37.6}    & {30.2}   \\
    \hline
    \multicolumn{3}{c}{\textit{Our LSM with Qwen2-7B}} \\
    {base}     & {45.3}    & {31.0}     \\
    {w/ AI-STA}     & \textbf{46.0}    & \textbf{31.4}   \\
    \bottomrule
  \end{tabular}
  \caption{Speech translation BLEU scores on CoVoST2. We conducted experiments in English (En)-to-Chinese (Zh), En-to-Japanese (Ja). For each result, We use \underline{underline} to highlight the previous SOTA baseline, and use \textbf{bold} to highlight surpassing the SOTA performance.}
  \label{tab:main_exp}
\end{table}

\textbf{Qwen-Audio}~\cite{chu2023qwen} is Alibaba's multi-modal LLM, accepting diverse audio and text inputs to output text. It proposes a multi-task learning framework and incorporates a word-level time-stamp prediction training task while yielding strong performance across various tasks. 

\textbf{Qwen2-Audio}~\cite{chu2024qwen2} is the latest progress of Qwen-Audio. It further boosts instruction-following capability and adopts direct preference optimization to align with human preferences achieving SOTA in AIR-Bench \cite{yang2024air}. 

\section{Results}
\subsection{Main Result}
Table~\ref{tab:main_exp} presents a comparison of our base LSM, our LSM with AI-STA method, and previous methods, reporting SacreBLEU scores evaluated on two language pairs: En-Zh, and En-Ja. Notably, our base LSM with Qwen2-7B achieves SOTA with the BLEU of 45.3 on En-Zh and 31.0 on En-Ja translation direction. Further with our AI-STA method, our LSM outperforms previous SOTA for 0.8 BLEU in En-Zh and 2.8 BLEU in En-Ja. Additionally, the AI-STA method provides a noticeable boost in performance for all models, with BLEU score gains of approximately 0.8 for LSM with Vicuna-13B-v1.1 and 0.6 for LSM with Qwen2-7B. We also observe that the performance gain with AI-STA is greater for En-Zh (0.9 BLEU) than for En-Ja (0.4 BLEU), suggesting that our alignment method may benefit more from target languages with richer training resources. These results convincingly demonstrate the superiority of AI-STA and highlight the promising potential of exploring LLMs for speech translation tasks.





\subsection{Effect of Optimization Layer Selection}\label{Aligning Layer Selection}
Tabel~\ref{tab:ablation_pos} shows that different aligning layer selections have a great impact on CoVoST2 En-Zh performance. Only aligning speech and text representation in layer 0 (embedding layer) gains 0.4 BLEU improvement. Once we further align at the inner layers, the performance begins to decline (45.7 -> 45.5). Especially when not aligning layer 0, the training loss fails to converge leading to catastrophic failure. The performance is further enhanced by aligning with selected alignment layers obtained through our optimization layer selection strategy (45.7 -> 46.0), highlighting the necessity of our layer selection strategy.

\begin{table}[!t]
  \centering
  \renewcommand\arraystretch{1.0}
  \begin{tabular}{
    p{0.2\textwidth}|
    >{\centering\arraybackslash}p{0.1\textwidth}}
    \toprule
    \textbf{Aligning Position} & \textbf{BLEU} \\
    \midrule
    {base}    & {45.3}      \\

    {Layer 0}   & {45.7}     \\

    {Layer 0-5}     & {45.5}     \\

    {Layer 1}   & {X}     \\

    {Layer 0-1(Selected)}   & \textbf{46.0}     \\
    \bottomrule
  \end{tabular}
  \caption{The impact of different layer optimization selection strategies on the performance of CoVoST2 En-Zh using Qwen2-7B as backbone LLM.}
  \label{tab:ablation_pos}
\end{table}

\subsection{Comparison of Aligning Methods}
For connectionist temporal classification (CTC) \cite{graves2006connectionist}, we apply it at the token level using backbone LLM's tokenizer to encode the transcript of source speech as the golden token and train an independent classification layer for matching with LLM's vocabulary size. For contrastive learning (CL), we treat golden paired speech-text samples as positive pairs, and the others as negative pairs and apply a multi-class N-pair contrastive loss \cite{sohn2016improved}. Both alignment methods only operate at the embedding layer. As shown in Table~\ref{tab:comparison_align_method}, employing either CTC or CL results in performance degradation. By contrast, our method yields a 0.7 BLEU improvement. 

We argue that CTC is not suitable for adapters like Q-Former that incorporate attention mechanisms. CTC is a forced alignment method where each output position must align precisely with a specific token, which can lead to conflicts when applied after the Q-Former. When applying contrastive learning (CL) such an alignment method at the overall semantic level, yields limited effectiveness and fails to further capture the fine-grained relationships between words. Both methods cause conflicting training objectives and hinder the training process.

\begin{figure}[t]
  \includegraphics[width=\columnwidth]{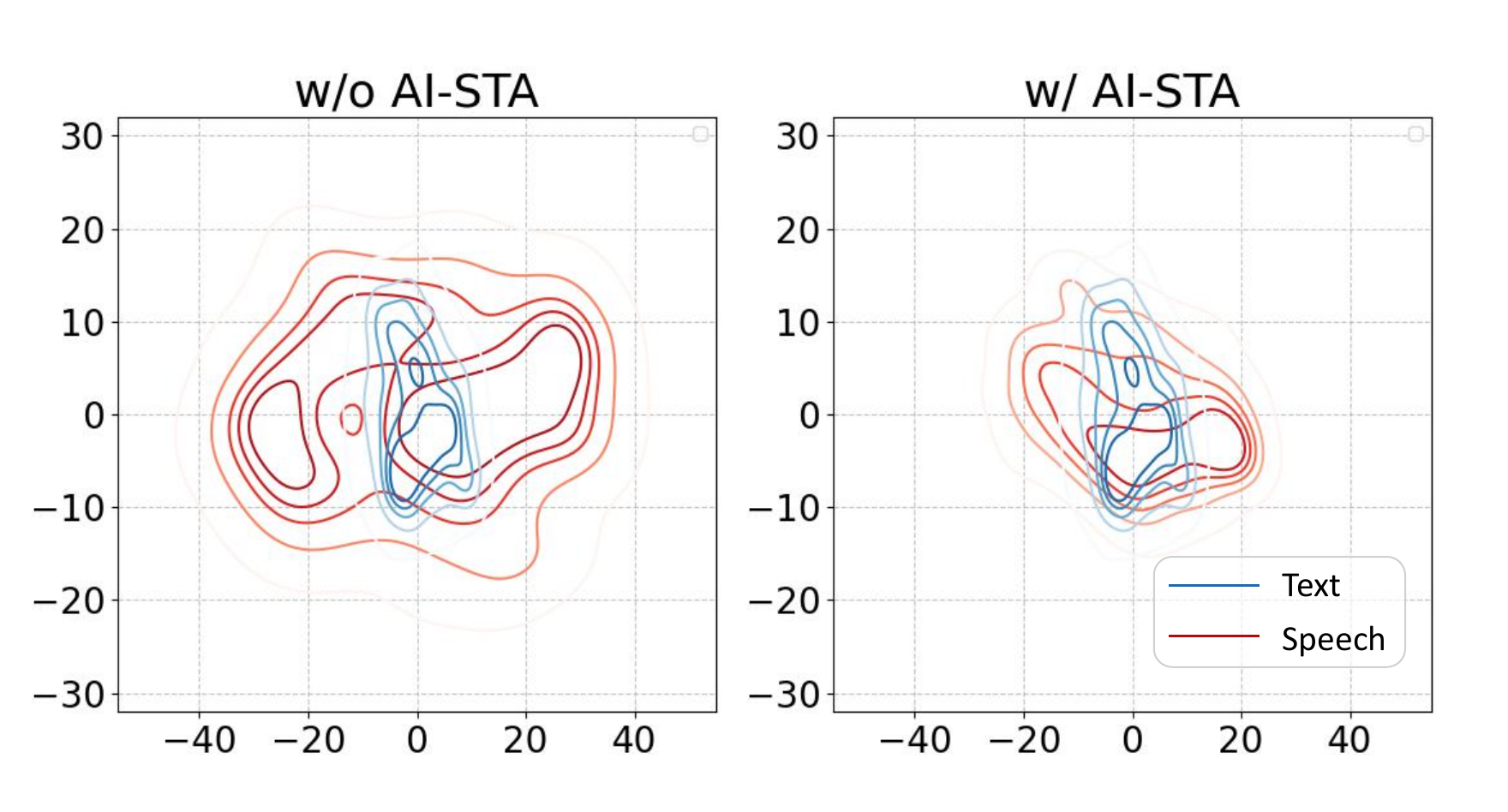}
  \caption{t-SNE visualization of speech and text representation from LSMs trained with or without AI-STA methods.}
  \label{fig:dimension_space}
\end{figure}

\begin{table}[!t]
  \centering
  \renewcommand\arraystretch{1.0}
  \scalebox{0.90}{
  \begin{tabular}{
    p{0.2\textwidth}|
    >{\centering\arraybackslash}p{0.1\textwidth}}
    \toprule
    \textbf{Aligning method} & \textbf{BLEU} \\
    \midrule
    {base}    & {45.3}    \\

    {$\hookrightarrow$ w/ CTC}   & {44.6}   \\

    {$\hookrightarrow$ w/ CL}   & {45.1}     \\

    {$\hookrightarrow$ w/ AI-STA}          & \textbf{46.0} \\
    \bottomrule
  \end{tabular}
  }
  \caption{The impact of different aligning methods on CoVoST2 En-Zh performance.}
  \label{tab:comparison_align_method}
\end{table}

\subsection{Can AI-STA Close the Modality Gap?}
We randomly sample 1,000 speech-text transcription pairs from the test set of CoVoST2 En-Zh to explore representation alignment between speech and text in the embedding layer of our LSM with the Qwen2-7B version. The speech representation is obtained as semantic tokens after processing through the Q-Former, while the text representation is derived from the tokenization and embedding layer of the LLM. All representations are averaged along the length dimension. We apply bivariate kernel density estimation \cite{parzen1962estimation} and utilize the T-SNE technique reducing data dimensions to a two-dimensional space for visualization \cite{van2008visualizing}. As depicted in Figure~\ref{fig:dimension_space}, AI-STA significantly reduces the distance discrepancy between the speech and text representation spaces compared to the baseline without our method, demonstrating a strong relationship between these two modalities.

\begin{figure}[t]
    \centering
  \includegraphics[width=0.9\columnwidth]{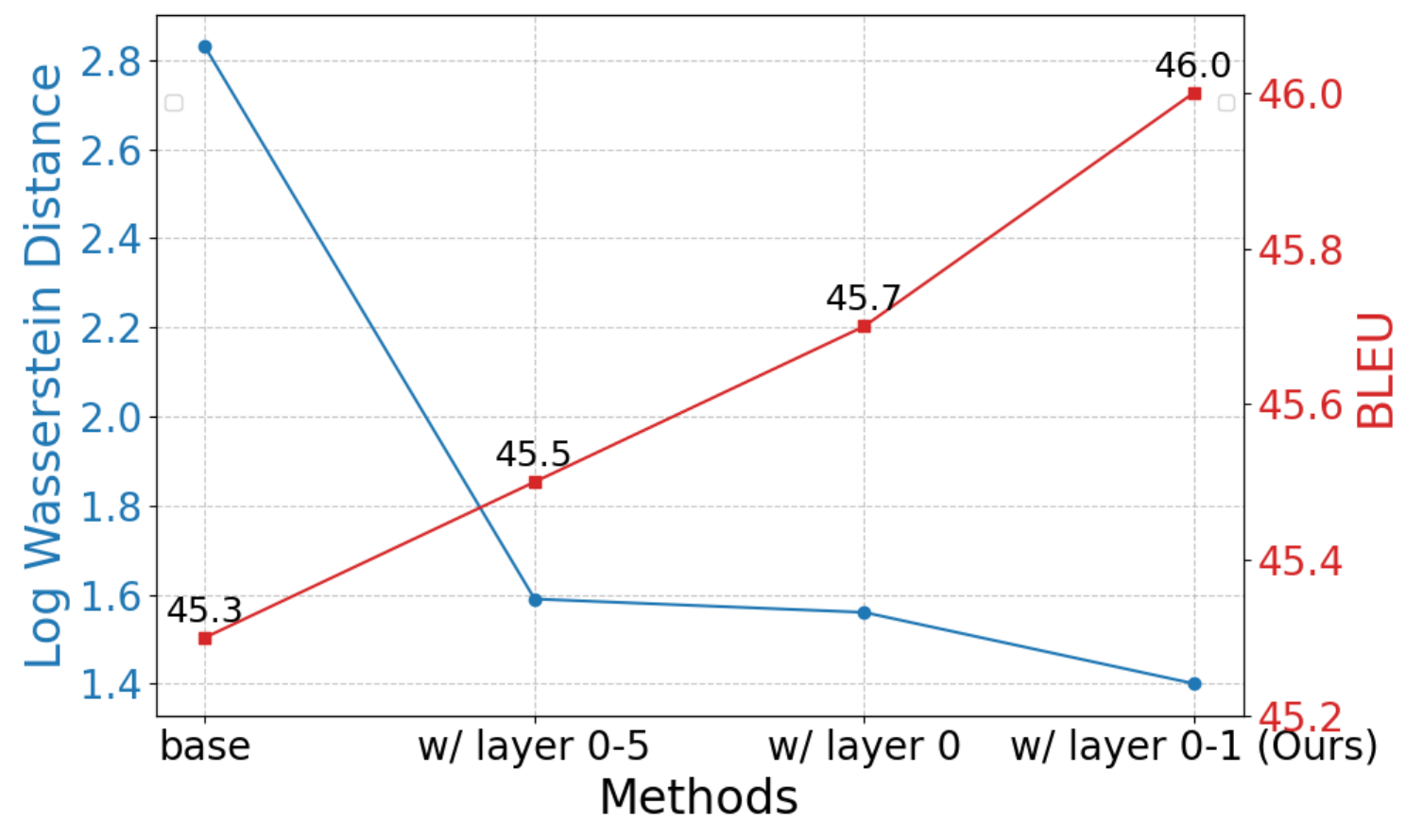}
  \caption{The correlation between alignment degree and performance in the CoVoST2 En-Zh direction across different training methods. Lower logarithmic Wasserstein distance indicates a higher degree of alignment, while a higher BLEU score corresponds to better performance in the ST task.}
  \label{fig:alignment_bLEU}
\end{figure}

\subsection{Correlation between Alignment Degree and ST Performance}
To investigate the relationship between representation alignment and speech translation performance, we calculate alignment scores for LSMs trained with various alignment methods. Taking the LSM with Qwen2-7B as an example, our optimal layer selection strategy identifies the zero and first layers as particularly suitable for representation alignment. We quantify the degree of alignment by calculating the logarithmic Wasserstein distance between these two layers. Figure~\ref{fig:alignment_bLEU} illustrates a strong correlation between the alignment score and speech translation performance. From left to right, the alignment methods are the same with section~\ref{Aligning Layer Selection} except for the optimization applied to layer 1. As the alignment scores decrease, we consistently observe a steady increase in the BLEU score, indicating a strong correlation between improved alignment and enhanced translation performance.

\subsection{Can AI-STA Help Knowledge Transfer?}
To investigate whether our method can bridge the modality gap and enable the model to understand speech modality inputs as text modality. We directly perform zero-shot text translation inference on a model that has been trained on the speech translation task. The inference prompt is identical to the training prompt. We intended to verify whether the model has effectively utilized the correspondence between speech and text during the training process. 

As shown in Table~\ref{tab:analysis_MT}, we observe that in the Vicuna-13B version, the zero-shot performance gap between using and not using our method reaches up to 5.4 BLEU. This indicates that our method significantly enhances the LLM's ability to leverage ST knowledge for text-based MT tasks. In the Qwen2-7B version, the zero-shot performance gap shrinks to 0.3 BLEU. Irrespective of whether our method is applied, the translation performance in text scenarios is significantly stronger than that in speech scenarios. We attribute this phenomenon to the Qwen2-7B model's strong English and Chinese language capabilities, as well as its more precise capture of the relationship between speech and text modalities. We use the transcript and translation text pair of the CoVoST2 En-Zh test set as our MT evaluation data.

\begin{table}[t]
  \centering
  \renewcommand\arraystretch{1.0}
  \scalebox{0.90}{
  \begin{tabular}{
    p{0.2\textwidth}|
    >{\centering\arraybackslash}p{0.1\textwidth}
    >{\centering\arraybackslash}p{0.1\textwidth}}
    \toprule
    \textbf{Method} & \textbf{ST} & \textbf{MT} \\
    \midrule
    \multicolumn{3}{c}{\textit{Training on ST}} \\
    {base(Vicuna-13B)}    & {36.5}  & {34.0}    \\

    {$\hookrightarrow$ w/ AI-STA}   & {37.6}  & {39.4}   \\
    { base(Qwen2-7B)}    & {45.3}  & {51.2}    \\
    {$\hookrightarrow$ w/ AI-STA}   & \textbf{46.0}  & \textbf{51.5}   \\
    \bottomrule
  \end{tabular}
  }
  \caption{The impact of Speech-Text Alignment on zero-shot machine translation task. Demonstrates that our method facilitates knowledge transferring from speech to text modality.}
  \label{tab:analysis_MT}
\end{table}


\subsection{Case Study}
\begin{figure}[t]
  \includegraphics[width=\columnwidth]{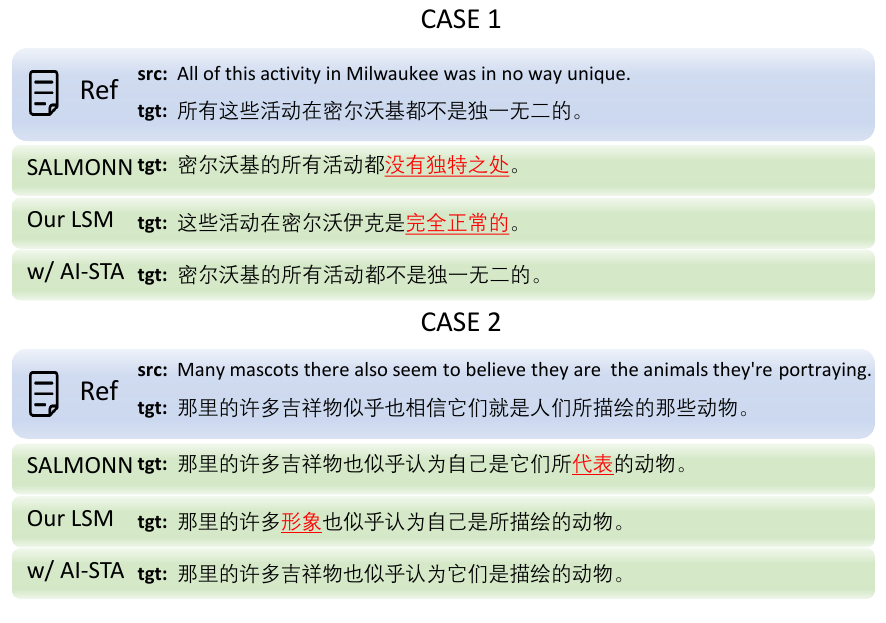}
  \caption{CoVoST2 En-Zh test cases that generated from the SALMONN, our LSM with Vicuna-13B and our LSM with AI-STA. The \textcolor{red}{\underline{red underlined text}} indicates an incorrect answer.}
  \label{fig:case_study}
\end{figure}

In this section, we present several cases generated by our LSM with Vicuna-13B to compare its performance with the previous end-to-end model, SALMONN \cite{tang2023salmonn}. The results are summarized in Figure~\ref{fig:case_study}. In the first case, SALMONN incorrectly translates ``in no way unique'' as the meaning of ``nothing unique'', leading to a deviation from the intended meaning. Our LSM inaccurately translates it as meaning ``completely normal'' which is out of context. While training with our AI-STA method, our LSM accurately translates it to the correct answer. 

In the second case, SALMONN and our LSM exhibit different translation errors in this case. SALMONN fails to translate the word ``portraying'' as ``representation''. Our LSM, in turn, misinterprets the word ``mascot'', resulting in a significant misunderstanding. In contrast, our LSM with AI-STA correctly translates these words, yielding a more accurate overall translation than SALMONN and our base LSM. These observations highlight the ST capabilities of our LSM with AI-STA, demonstrating that AI-STA enhances the LSM's ability to capture fine-grained semantics.

Although our method improves the performance of LLM-based ST, we still discovered content omission during the translation generation process in the two cases mentioned above. Some source words remain untranslated, such as "all of this" in the first case, which is not fully translated by any of the methods. As a result, the translated sentences tend to be shorter than the reference sentences. This highlights a persistent issue in current LLM-based Speech Translation systems, suggesting that there is still room for improvement.

\section{Conclusions}
In this study, we enhance the ST capabilities of LSMs by explicitly aligning speech and text representations. To achieve this, we introduce OT theory to quantify the discrepancy between speech and text representations and investigate the representation characteristics of different layers within LLMs. By leveraging the cross-modal retrieval technique, we identify specific model layers that are well-suited for representation alignment and perform joint training using these selected layers. Our experiments demonstrate that this method effectively reduces the distance between the speech and text representation spaces, enabling the model to better capture the relationships between the two modalities and significantly improves the performance of large speech models on speech translation tasks.
\section*{Limitation}
We acknowledge that our proposed approach has several limitations: (1) We observed several intriguing phenomena, such as performance degradation when applying CTC or CL alignment methods at the embedding layer, as well as a sharp drop in retrieval performance at certain layers within LLM and remained low in subsequent layers. However, we did not thoroughly investigate the underlying principles and instead relied on intuition and empirical observations without theoretical justification or formal proof. (2) Although our method enhances the LLM-based ST performance and reaches SOTA, a performance gap remains compared to the text scenarios' machine translation. However, we believe our method provides valuable insights and encourages the development of cross-modal learning in LLMs.

\bibliography{custom}

\appendix

\section{Appendix}
\label{sec:appendix}

\subsection{Hyperparameters}\label{sec:Hyperparameters}
The training configurations for the two stages are summarized as follows:\\
\textbf{Speech Pre-training Stage}:~Training employs the AdamW optimizer with hyperparameters $\beta_{1}$=0.9, $\beta_{2}$=0.999, $eps=1e^{-8}$. The learning rate follows a cosine decay schedule, starting with a warm-up rate of $1e^{-6}$, peaking at $3e^{-5}$, and decaying to a minimum of $1e^{-5}$. Weight decay is set to 0.05, and the global batch size is 32. The model undergoes 80k training steps with 9k warm-up steps, using \textsc{bfloat16} numerical precision. LoRA parameters include a rank of 8, alpha of 32, and dropout of 0.1.\\
\textbf{Joint Training Stage}:~The training configuration for this stage is largely the same as the mentioned above, with two differences: the warm-up steps in this stage are set to 3k, and we do not fix the total number of training steps. Instead, we determine whether to stop training based on the metrics from the validation phase conducted every 3k training steps. Training is stopped when the validation accuracy does not exceed the previous highest value for four consecutive validation phases.


\end{document}